\documentclass{article}
\usepackage{spconf,amsmath,graphicx}
\usepackage{booktabs} 
\usepackage{multirow} 
\usepackage{amsfonts}
\usepackage{url}
\usepackage{enumitem}
\usepackage{color}
\usepackage[table]{xcolor}
\usepackage{mwe} 
\newcommand{\mypar}[1]{\vspace{0.15em}
\textbf{#1}~}

\usepackage{bbm}
\title{Semi-Supervised Medical Image Segmentation via Dual Networks}

\name{Yunyao Lu$^1$, Yihang Wu$^1$, Reem Kateb$^2$, Ahmad Chaddad$^{1,3,*}$ }

\address{\parbox{1 \linewidth}{\centering $^1$AIPM, School of Artificial Intelligence, Guilin University of Electronic Technology, China\\
$^2$College of Computer Science and Engineering, Jeddah University, Jeddah, Saudi Arabia.\\
$^3$Laboratory for Imagery, Vision and Artificial Intelligence, École de Technologie Supérieure, Canada\\ Correspondence:ahmad8chaddad@gmail.com
}}
\begin{document}
%
\maketitle
\begin{abstract}
Traditional supervised medical image segmentation models require large amounts of labeled data for training; however, obtaining such large-scale labeled datasets in the real world is extremely challenging. Recent semi-supervised segmentation models also suffer from noisy pseudo-label issue and limited supervision in feature space. To solve these challenges, we propose an innovative semi-supervised 3D medical image segmentation method to reduce the dependency on large, expert-labeled datasets. Furthermore, we introduce a dual-network architecture to address the limitations of existing methods in using contextual information and generating reliable pseudo-labels. In addition, a self-supervised contrastive learning strategy is used to enhance the representation of the network and reduce prediction uncertainty by distinguishing between reliable and unreliable predictions. Experiments on clinical magnetic resonance imaging demonstrate that our approach outperforms state-of-the-art techniques. Our code is available at \url{https://github.com/AIPMLab/Semi-supervised-Segmentation}. 
\end{abstract}


\section{Introduction}
Medical image segmentation is a fundamental technique designed to partition images into distinct regions, playing an important role in the medical field. The identification and quantification of abnormal regions (e.g., abnormal areas) provide clinicians with critical diagnostic and therapeutic insights, enabling more informed medical decision making \cite{wang2022medical}. Many deep learning architectures \cite{ronneberger2015u,cciccek20163d,isensee2021nnu} have shown promising results, achieving state-of-the-art (SOTA) performance in various medical image segmentation tasks \cite{jiao2023learning}.

However, traditional supervised learning methods are based on large-scale labeled datasets, which are not practical in the medical domain. Furthermore, pixel-level annotations in medical images are particularly challenging and labor intensive \cite{wang2023dual}. To address the impact of limited labeled data, in \cite{jiao2023learning}, they summarized semi-supervised learning (SSL) techniques, which can leverage unlabeled data with labeled data for model training to reduce dependence on large labeled datasets. Due to the unavailability of ground truth for unlabeled samples, pseudo-supervision is proposed for SSL. The key idea is to use the pseudo-labels predicted by the segmentation model for model training \cite{yang2022survey}. 

Despite advances in pseudo-labeling methods, two major challenges remain in the task of semi-supervised medical image segmentation: 1) \textit{Noisy pseudo-label}, and 2) \textit{insufficient supervision in the feature space}. Pseudo-supervision relies on segmentation models to generate pseudo-labels for unlabeled images. However, these models are highly susceptible to label noise, which weakens the effectiveness of the pseudo labeling process. Similarly, insufficient supervision in the feature space also limits the performance of SSL \cite{zhao2023rcps}. Many methods only provide supervision in the label space and lack explicit guidance in the feature space, resulting in insufficient class separability \cite{ouali2020semi, chen2021semi, tarvainen2017mean, zhang2021flexmatch}. Although recent studies have attempted to mitigate noise by filtering out predictions with classification scores below a certain threshold \cite{zhang2021flexmatch}, this approach does not fully eliminate false predictions (e.g., some incorrect predictions may still yield high classification scores, leading to overconfidence or miscalibration \cite{guo2017calibration}). Furthermore, using a high threshold to exclude unreliable predictions considerably decreases the number of pseudo labels, which in turn impacts the effectiveness of SSL. This reduction in pseudo-labels can lead to class imbalance in the training data, reducing overall segmentation performance \cite{karimijafarbigloo2024leveraging}.

Motivated by the previous challenges, this study proposes a dual-stream network architecture inspired by the idea of Knowledge Distillation (KD) with S-T frameworks, where each sub-network has a 3D encoder-decoder module. Furthermore, we introduce a supervised loss function to guide the network in learning the features of each class. Additionally, to stabilize the training process, we involve consistency regularization by minimizing the cosine distance between the predictions of the student and teacher networks. The teacher network generates pseudo labels for unlabeled data, while the student network uses those data as supervisory features for learning. In addition, to alleviate the negative impact of noisy pseudo-labels on model performance, we dynamically adjust the contribution of pseudo-labels to model training based on uncertainty estimation. Specifically, uncertainty estimation introduces adaptive voxel weighting for pseudo-supervised loss, where voxels with higher confidence are assigned higher weights, and vice versa. Finally, we use contrastive learning to align the uncertain pixel features with the prototypes of reliably predicted classes, while increasing the distance between different class prototypes. Our contributions can be summarized as follows. 
\begin{enumerate}
     \item We introduce a consistency regularization term to reduce discrepancies between student and teacher network predictions. It leads to stabilize the training process.
    \item We propose an uncertainty estimation mechanism to dynamically adjust pseudo-label contributions, mitigating noise and avoiding over depended on uncertain regions.
    \item We use contrastive loss to align unreliable pixel features with reliable class prototypes, while increasing separation between different class prototypes.
\end{enumerate}

\section{PROPOSED METHOD}
\begin{figure}
    \centering
    \includegraphics[width=1\linewidth]{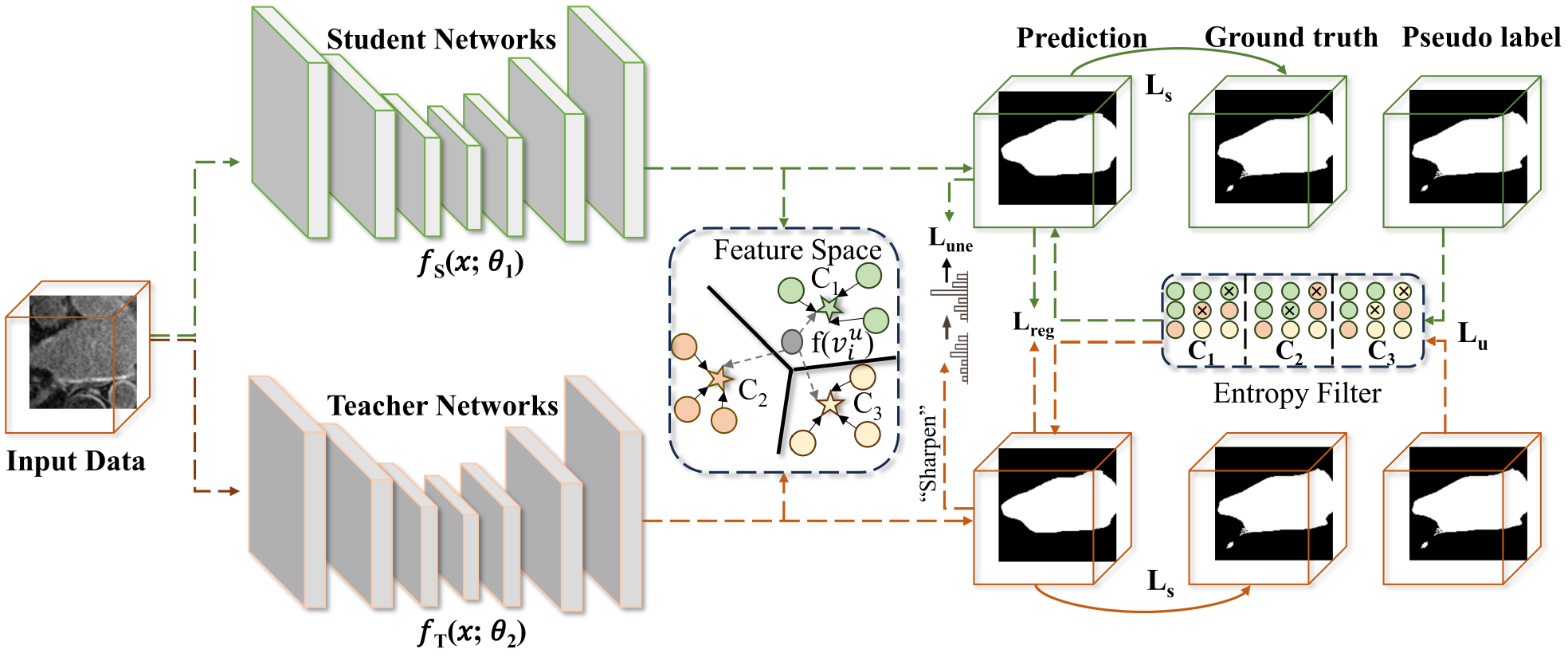}
    \caption{Pipeline of our proposed model. It has three parts: consistency regularization, uncertainty estimation, and a contrastive learning method to improve student and teacher performance. }
    \label{fig:model}
\end{figure}




Figure \ref{fig:model} shows the pipeline of our proposed method. It consists of two subnetworks: the Student network \( f_S(x; \theta_1) \) and the Teacher network \( f_T(x; \theta_2) \)
both employing a 3D encoder-decoder design with comparable performance. It uses a combination of labeled data \(
D_l = \left\{ \left( x_i^l, y_i^l \right) \right\}_{i=1}^{N_l}
\) and a larger set of unlabeled data \(
D_u = \left\{ x_i^u \right\}_{i=1}^{N_u}
\) to enhance the segmentation performance. Furthermore, we introduce consistency regularization, uncertainty estimation, and contrastive loss to improve the performance of the student and teacher model. 

In general, the total loss is divided into supervision loss $\mathcal{L}_s$, self-supervised contrastive loss $\mathcal{L}_c$, semi-supervised pseudo-labeling loss $\mathcal{L}_u$, and regularization loss $\mathcal{L}_{reg}$.

\begin{equation}
\mathcal{L} = \underbrace{\mathcal{L}_s^{S} + \mathcal{L}_s^{T}}_{\text{labelled}} 
+  \lambda_c \underbrace{\mathcal{L}_{c} + \mathcal{L}_{reg} + \mathcal{L}_{une}}_{\text{self}} 
+ \underbrace{\mathcal{L}_u^{S} + \mathcal{L}_u^{T}}_{\text{semi}}
\label{eq:overall_loss}
\end{equation}

For the labeled data, our aim is to minimize the Cross-Entropy loss and Dice loss, as defined by the following Equation \ref{eq:supervised_loss}. 

\begin{equation}
\mathcal{L}_s = \mathcal{L}_{ce}(p_i^l, y_i^l) + \text{Dice}(\sigma(p_i^l), y_i^l)
\label{eq:supervised_loss}
\end{equation}

Here, \( y_i^l \) represents the manually annotated mask label for the \( i \)-th labeled image, while \( p \) denotes the model prediction. 

For the unlabeled data, we apply pixel-level entropy filtering to remove unreliable pseudo-labels prior to computing the unsupervised loss. We introduce an unsupervised loss function based on entropy minimization, aimed at reducing the uncertainty in model predictions and preventing the model from being misled by identifying and excluding areas with high uncertainty. Specifically, we calculate the entropy of each pixel to assess the confidence of the model in its predictions. Regions with the greatest uncertainty are filtered out according to a predefined ratio and assigned a special ignore label. Subsequently, we employ a cross-entropy loss function with an ignored index to compute the loss, thereby effectively minimizing the influence of noise on the model.

\begin{equation}
{Ent}_S = - \sum_{c=1}^{C} {p}_i^u(c) \cdot \log_2 \left( {p}_i^u(c) + \epsilon \right),
\end{equation}
where \({p}_i^u(c) \) represents the predicted probability for class \(c\). \({Ent}_S\) denotes the entropy value of the student model, and \({Ent}_T\) count in the same way.

Define a mask for valid points based on a threshold \(\tau\), while ignoring high-entropy pixels:
\[
\hat{y}^u_{i} =
\begin{cases}
\arg\max_c \, p_{i}^u(c), & \text{if } {Ent} < \tau, \\
\text{ignore}, & \text{otherwise.}
\end{cases}
\]
where \(\tau\) represents the 80th percentile of the entropy values across all pixels.

We calculate the unsupervised loss defined in Equation \ref{eq:unsupervised_loss}:
\begin{equation}
\mathcal{L}_{u} = \frac{\sum_{i=1}^{B \times H \times W \times D} \mathcal{L}_{ce}(f(x^u; \theta), \mathbb{I}\left[ \hat{y}^u_i \neq \text{ignore} \right]) }{\sum_{i=1}^{B \times H \times W \times D} \mathbb{I}\left[ \hat{y}^u_i \neq \text{ignore} \right]},
\label{eq:unsupervised_loss}
\end{equation}
where \( B, H, W, D \) represent the batch size, height, width, and depth of the input data, respectively. The function \( \mathbb{I}(\cdot) \) is an indicator function.

\subsection{Consistency Regularization}
Relying on the smoothness assumption, where small perturbations to an input should not result in significantly different predictions \cite{han2024deep}, model consistency becomes a crucial objective. Specifically, model consistency demands that when the same image is passed through different models, the predictions should be similar. However, using pseudo-labels can lead to increased risk of training instability. To address this, we introduce a consistency regularization term that enforces prior constraints based on these assumptions, ensuring more stable training.

To further encourage consistency between the two subnetworks $S$ and $T$, we employ a cosine similarity-based consistency regularization. This regularization aims to minimize the cosine distance between the predictions of the subnets $S$ and $T$, reducing their disagreement, and thus stabilizing the training. The consistency regularization loss is defined as:

\begin{equation}
\mathcal{L}_{reg} = 1 - \cos(p_i^S, p_i^T)
\end{equation}

Where \( p_i^S \) refers to the \( i \)-th output of the student model, and \( p_i^T \) represents the \( i \)-th output of the teacher model. The cosine similarity between the outputs of the two subnetworks is calculated as:

\begin{equation}
\cos(p_i^S, p_i^T) = \frac{p_i^S \cdot p_i^T}{\|p_i^S\| \|p_i^T\|}
\end{equation}

By minimizing this loss, we ensure that the predictions from different models of the same input remain consistent, thus reducing variance and improving model generalization.

\subsection{Uncertainty Estimation}
Pseudo supervision can sometimes be unreliable due to label noise. A common method to reduce this noise is confidence threshold, but this approach can fail in segmentation tasks, as it may bias the model towards easier classes, leaving harder classes underrepresented. To address this challenge, we propose using prediction uncertainty to refine pseudo supervision. By estimating uncertainty through KL-divergence, we adjust the contribution of the pseudo labels based on prediction confidence.

The uncertainty loss is defined as:
\begin{equation}
\mathcal{L}_{p}(p_i^S, p_i^T) = \mathcal{L}_{ce}(p_i^S, \sigma(p_i^T / Tp)),
\end{equation}
\begin{equation}
    \mathcal{L}_{une}(p_i^S, p_i^T) = e^{-\mathcal{D}_{KL}(p_i^S, p_i^T} )\mathcal{L}_{p}(p_i^S, p_i^T) + \mathcal{D}_{KL}(p_i^S, p_i^T)
\end{equation}
\begin{equation}
D_{KL}(p_i^S, p_i^T) = p_i^T \log \left(\frac{p_i^T}{p_i^S}\right).
\end{equation}
where, $\mathcal{D}_{KL}(p_i^S, p_i^T)$ represents the KL-divergence between the model prediction and the pseudo label. We sharpen the softmax probabilities by dividing the logits by a temperature hyperparameter $T_p$.

\subsection{Contrastive Loss for Uncertain Voxels}
To address the issue of  insufficient supervision in the feature space, we incorporate a contrastive loss function. The goal is to align uncertain voxels with their respective class prototypes, reducing the risk of misclassification and minimizing prediction uncertainty. This approach estimates the confidence level of each voxel prediction and classifies the predictions into two categories: reliable and unreliable. Based on the reliable predictions, class prototypes are constructed by averaging the representations of these reliable voxels. 

Initially, we calculate the foreground and background prototypes, which are the mean vectors of the reliable voxel representations. These prototypes represent the class identity within the feature space. The foreground class prototype is calculated as:

\begin{equation}
    c_f = \frac{1}{|S_f|} \sum_{(\mathbf{v}_i^r, y_i) \in S_f} f(\mathbf{v}_i^r)
\end{equation}

The calculation for the background class prototype, \( c_b \), follows the same equation.

Where \( S_f \) denotes the sets of reliable voxels for the foreground and background classes, respectively. \( f(\mathbf{v}_i^r) \) represents the feature embedding of a reliable voxel \( \mathbf{v}_i^r \).

The contrastive loss function is composed of three terms, designed to align uncertain voxels with their corresponding class prototypes while maximizing the separation between different prototypes:

\begin{equation}
    \mathcal{L}_{c} = d(f(\mathbf{v}_i^u), c_f) + d(f(\mathbf{v}_i^u), c_b) + d(c_f, c_b)
\end{equation}

The function \( d(\cdot, \cdot) \) calculates the distance between the voxel’s feature embedding and the class prototypes, encouraging the alignment of uncertain voxels to the most appropriate class prototype.

\vspace{0.4 cm}
\begin{table}[!ht]\scriptsize
\renewcommand{\arraystretch}{0.8}
\setlength{\tabcolsep}{4.6pt}
\caption{Summary of test metrics in LA dataset (MRI images). The best two results are marked in bold.}
\label{LA:1} 
\begin{tabular}{ccccc}
\hline
\textbf{Method} & \textbf{Dice (\%)} & \textbf{Jaccard (\%)} & \textbf{95HD (voxel)} & \textbf{ASD (voxel)} \\
\hline
MT \cite{tarvainen2017mean}     & 85.89 ± 0.024       & 76.58 ± 0.027        & 12.63 ± 5.741         & 3.44 ± 1.382          \\
UA-MT \cite{yu2019uncertainty}   & 85.98 ± 0.014       & 76.65 ± 0.017        & 9.86 ± 2.707          & 2.68 ± 0.776          \\
SASSNet \cite{li2020shape} & 86.21 ± 0.023       & 77.15 ± 0.024        & 9.80 ± 1.842          & 2.68 ± 0.416          \\
DTC \cite{luo2021semi}    & 86.36 ± 0.023       & 77.25 ± 0.020        & 9.92 ± 1.015          & 2.40 ± 0.223          \\
MC-Net \cite{wu2021semi} & 87.65 ± 0.011       & 78.63 ± 0.013        & 9.70 ± 2.361          & 3.01 ± 0.700          \\
MCF \cite{wang2023mcf}    & 88.71 ± 0.018       & 80.41 ± 0.022        & 6.32 ± 0.800         & 1.90 ± 0.187          \\
LUSEG \cite{karimijafarbigloo2024leveraging} & 89.10 ± 0.012 & 81.62 ± 0.024 & 6.30 ± 0.850 & 1.80 ± 0.020 \\
\rowcolor{gray!15} Ours &\textbf{90.19 ± 0.012} &\textbf{82.42  ±  0.019}&\textbf{5.74 ± 0.757} &\textbf{1.75 ± 0.227}  \\
\hline
\end{tabular}
\end{table}

\begin{table}[ht]\scriptsize
\renewcommand{\arraystretch}{0.8}
\setlength{\tabcolsep}{4.6pt}
\caption{Summary of test metrics in Pancreas dataset (CT images).}
\label{Pancreas:1}
\begin{tabular}{ccccc}
\hline
\textbf{Method} & \textbf{Dice (\%)} & \textbf{Jaccard (\%)} & \textbf{95HD (voxel)} & \textbf{ASD (voxel)} \\
\hline
MT \cite{tarvainen2017mean}      & 74.43 ± 0.024       & 60.53 ± 0.030        & 14.93 ± 2.000         & 4.61 ± 0.929          \\
UA-MT \cite{yu2019uncertainty}   & 74.01 ± 0.029       & 60.00 ± 3.031        & 17.00 ± 3.031         & 5.19 ± 1.267          \\
SASSNet \cite{li2020shape} & 73.57 ± 0.017       & 59.71 ± 0.020        & 13.87 ± 1.079         & 3.53 ± 1.416          \\
DTC \cite{luo2021semi}     & 73.23 ± 0.024       & 59.18 ± 0.027        & 13.20 ± 2.241         & 3.83 ± 0.925          \\
MC-Net \cite{wu2021semi}  & 73.73 ± 0.019       & 59.19 ± 0.021        & 13.65 ± 3.902         & 3.92 ± 1.055          \\
MCF \cite{wang2023mcf}     & 75.00 ± 0.026       & 61.27 ± 0.030        & 11.59 ± 1.611         & 3.27 ± 0.919          \\
LUSEG \cite{karimijafarbigloo2024leveraging} & \textbf{76.40 ± 0.018} & \textbf{62.96 ± 0.027} & \textbf{10.69 ± 1.603}  & 2.79 ± 0.0954 \\
\rowcolor{gray!15} Ours & 75.25 ± 0.014 & 61.65 ± 0.017 & 11.15 ± 1.604 & \textbf{2.49 ± 0.293 }\\
\hline
\end{tabular}
\end{table}

\begin{table*}[!ht]
\centering
\renewcommand{\arraystretch}{1}
\caption{Ablation study with loss functions.}
\label{Ablation}
\small
\resizebox{\textwidth}{!}
{\begin{tabular}{c|c|c|c|c|c|c|c|c|c}
\hline
\multirow{2}{*}{\centering $\mathcal{L}_{une}$} & \multirow{2}{*}{\centering $\mathcal{L}_{reg}$} & \multicolumn{4}{c|}{Left Atrium} & \multicolumn{4}{c}{Pancreas-CT} \\
\cline{3-10}
& & Dice (\%)  & Jaccard (\%)  & HD95 (voxel)  & ASD (voxel)  & Dice (\%)  & Jaccard (\%)  & HD95 (voxel) & ASD (voxel) \\
\hline
\checkmark & & 88.84 ± 0.011 & 80.38 ± 0.015 & 6.76 ± 0.802 & 2.16 ± 0.292 &\textbf{75.44 ± 0.018}&\textbf{61.80 ± 0.021 }&11.37 ± 1.70 &3.17 ± 0.128  \\
& \checkmark & 89.28 ± 0.012 & 81.02 ± 0.017 & 6.58 ± 0.628 & 2.06 ± 0.269 &73.92 ± 0.021 &60.10 ± 0.025&\textbf{10.13 ± 1.47} &\textbf{2.17 ± 0.248 }\\
\checkmark & \checkmark &\textbf{90.19 ± 0.012} &\textbf{82.42  ±  0.019}&\textbf{5.74 ± 0.757} &\textbf{1.75 ± 0.227} & 75.25 ± 0.014 & 61.65 ± 0.017 &11.15 ± 1.60 & 2.49 ± 0.293 \\
 &  & 89.03 ± 0.012 &80.61 ± 0.016 &6.68 ± 0.759 &2.24 ± 0.294 &74.62 ± 0.023  &60.83 ± 0.027 &11.17 ± 0.95 &2.81 ± 0.302 \\
 
\hline
\end{tabular}}
\end{table*}

\section{Experiments}

\subsection{Datasets}
\textit{The Left Atrial Dataset (LA).} The LA \cite{xiong2021global} consists of 100 3D gadolinium-enhanced MR volumes with manual left atrial annotations at an isotropic resolution of 0.625$\times$0.625 $\times$0.625 mm³. The dataset is divided into 5 folds of 20 volumes each. Preprocessing includes normalization to zero mean and unit variance according to \cite{wang2023mcf, karimijafarbigloo2024leveraging}. During training, random cropping generates input volumes of size 112 × 112 × 80. For inference, a sliding window approach of the same dimensions and a stride of 18$\times$18$\times$4 is used. \textit{The Pancreas Dataset.} These datasets include 82 contrast-enhanced 3D CT volumes with manual pancreas annotations \cite{roth2015deeporgan}. Each volume size is 512$\times$512$\times$D, where D ranges from 181 to 466 slices. The dataset is split into four folds. Preprocessing involves applying a soft tissue window (-120 to 240 HU) and cropping around the pancreas with a 25-voxel margin according to \cite{wang2023mcf, karimijafarbigloo2024leveraging}. Training uses random cropping to resize volumes to 96$\times$96$\times$6, and inference is performed with a sliding window of 16$\times$16$\times$16.

\mypar{Implementation details.}
Following \cite{wang2023mcf, karimijafarbigloo2024leveraging}, we adopted ResNet and V-Net as the two-stream network backbone, and used K-fold cross-validation (i.e., 5 folds for LA and 4 folds for Pancreas) to provide a comprehensive evaluation. The SGD optimizer (weight decay: 0.0001, momentum: 0.9) is used to optimize the network parameters, starting with a learning rate of 0.01, which decreased by a factor of 10 every 2500 iterations for a total of 6000 iterations. Each training iteration included both labeled and unlabeled samples, with a batch size of two. In Eq. \ref{eq:overall_loss}, we set \( \lambda_c = 0.1 \times e^{4(1 - \frac{t}{t_{\text{max}}})^2} \), where \( t \) and \( t_{\text{max}} \) represent the current and maximum iterations, respectively. The experiment environment is based on the Windows 11 operating system and features an Intel 13900KF CPU with 128 GB of RAM and an L40 GPU. We considered the Dice score, Jaccard, Hausdorff distance (HD95), and Average Surface Distance (ASD) to compare the performance of the models \cite{muller2022towards}.

\begin{figure}
    \centering
    \includegraphics[width=1\linewidth]{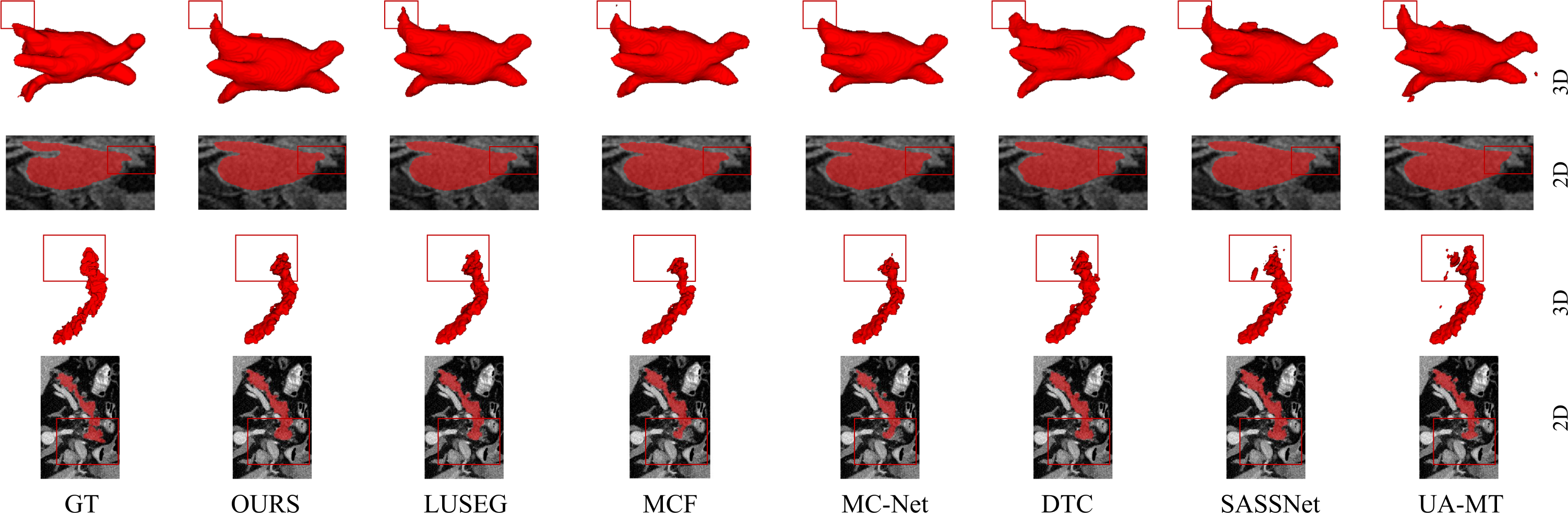}
    \caption{Visual comparison of segmentation results: the first and second rows show the left atrium (LA), while the third and fourth rows show the pancreas, respectively.}
    \label{fig:fig}
\end{figure}

\mypar{Results.} Table \ref{LA:1} and Table \ref{Pancreas:1} report the performance metrics using the LA and the Pancreas dataset. In Table \ref{LA:1}, compared to LUSEG, our model shows an improvement in Dice, with Dice increasing from 89.10 to 90.19. In addition, our method maintains a low performance variance, which improves the stability and reliability. Table \ref{Pancreas:1} demonstrates that our model achieves feasible performance in ASD. Figure \ref{fig:fig} illustrates the visual results of the left atrial and pancreas datasets on different segmentation methods. Within the pancreas dataset, our model shows lower error rates than the MCF and LUSEG models.

\mypar{Ablation study.}
Table \ref{Ablation} presents the effects of the regularization and uncertainty estimation elements within our model. In particular, these two losses ($\mathcal{L}_{une}$ and $\mathcal{L}_{reg}$) have different effects on the two datasets. For LA dataset, removing the regularization module resulted in a drop in the Dice from 90.19 to 88.84. Similarly, removing the uncertainty loss resulted in a drop in the Dice score from 90.19 to 89.28. Dice score decreases from 90.19 to 89.03 when both losses are absent.  

However, in pancreas datasets, we observed that the combination of both losses ($\mathcal{L}_{reg}$ and $\mathcal{L}_{une}$) resulted in lower Dice and Jaccard scores compared to using only uncertainty loss ($\mathcal{L}_{une}$). Yet, the combination led to better 95HD and ASD scores than using only the regularization loss ($\mathcal{L}_{reg}$). This result may be assigned to the varying features of the two datasets. For example, the Pancreas-CT dataset contains more complex structures and varied image features, which may cause the combination of losses to overfit or struggle to generalize well. The additional complexity of combining both losses might lead to better boundary (HD and ASD) performance, but at the cost of a reduction in overlap metrics (Dice and Jaccard). This suggests that for some complex datasets, focusing on individual loss components may yield better results, while the combined effect could introduce noise that affects segmentation performance.

\section{CONCLUSION}
In this paper, we presented a 3D network, an innovative contrastive consistency segmentation model designed to effectively leverage limited annotations for medical image segmentation. By incorporating an enhanced contrastive learning strategy alongside consistency and uncertainty estimation schemes, our approach achieves promising results on CT and MRI images. In future work, we will involve domain adaptation \cite{10835760} to solve domain shifts that exist among different modalities to further boost the performance of our model in multi-source situations.

\section{Compliance with ethical standards}
\label{sec:ethics}
This is a numerical simulation study for which no ethical approval was required.

\vspace{-3pt}
\section{Acknowledgements}
\vspace{-3pt}
This research was funded by the National Natural Science Foundation of China \#82260360, the Guilin Innovation Platform and Talent Program \#20222C264164, and the Guangxi Science and Technology Base and Talent Project (\#2022AC18004 and \#2022AC21040).

\footnotesize
\bibliographystyle{ieeetr}
\bibliography{refs}

\end{document}